\documentclass[10pt,twocolumn,letterpaper]{article}

\usepackage[pagenumbers]{iccv} 

\definecolor{iccvblue}{rgb}{0.21,0.49,0.74}
\usepackage[pagebackref,breaklinks,colorlinks,allcolors=iccvblue]{hyperref}

\title{VRMDiff: Text-Guided Video Referring Matting Generation of Diffusion}

\author{
Lehan Yang$^{1}$~~~
Jincen Song$^{2}$~~~
Tianlong Wang$^{1}$~~~ \\
Daiqing Qi$^1$~~~  
Weili Shi$^1$ ~~~
Yuheng Liu$^{3}$ ~~~
Sheng Li$^{1}$ ~~~
\\[0.2cm]
$^1$University of Virginia~~
$^2$Columbia University~~ 
$^3$Texas A\&M University~~ 
}

\newcommand{\ourmethod}{VRMDiff}
\newcommand{\ourdataset}{VRM-10K}
\newcommand{\ourloss}{Latent-Constructive}

\usepackage{pifont} 

\newcommand{\cmark}{\ding{51}} 
\newcommand{\xmark}{\ding{55}}

\begin{document}
\maketitle
\begin{abstract}

We propose a new task, video referring matting, which obtains the alpha matte of a specified instance by inputting a referring caption. 
We treat the dense prediction task of matting as video generation, leveraging the text-to-video alignment prior of video diffusion models to generate alpha mattes that are temporally coherent and closely related to the corresponding semantic instances. 
Moreover, we propose a new~\ourloss{} loss to further distinguish different instances, enabling more controllable interactive matting. 
Additionally, we introduce a large-scale video referring matting dataset with 10,000 videos. To the best of our knowledge, this is the first dataset that concurrently contains captions, videos, and instance-level alpha mattes.
Extensive experiments demonstrate the effectiveness of our method.
The dataset and code are available at \url{https://github.com/Hansxsourse/VRMDiff}.
 
\end{abstract}    
\section{Introduction}
\label{sec:intro}

Video matting is the process of extracting precise foreground object boundaries from video sequences, widely used in video editing, film production, and augmented reality. 
Compared to video segmentation~\cite{botach2022end,cheng2023tracking,li2024univs,yuan2024losh,cho2024dual,ravi2025sam}, matting provides transparency information, better capturing edges and the effects of lighting. 
Previous video matting techniques~\cite{zou2015video,wang2021video,lin2022robust,park2023mask} typically focus on extracting the foreground without distinguishing between different instances. 
Recently, some works have differentiated between instances but cannot provide their semantic labels~\cite{li2024video,huynh2024maggie}, while they rely on the guidance of masks. 
Although these methods achieve impressive results, their scalability and usability are limited, especially in scenes involving multiple objects or when specific objects need to be designated.

Referring image matting~\cite{li2023referring,li2024matting} has emerged as a potential solution to these limitations, utilizing natural language expressions to specify target objects in images. 
This approach allows for intuitive selection of target objects without needing post-adjusting, enhancing flexibility and user experience. 
However, extending referring image matting to videos presents significant challenges. 
Videos contain temporal dynamics, object motion, and complex interactions between multiple objects, requiring models capable of understanding spatiotemporal information and maintaining temporal consistency across frames. 
Directly applying image referring matting methods to videos lacks temporal information, making it challenging to maintain the consistency of the same instance across different frames.

\begin{figure}[tp]
    \centering
    \includegraphics[width=0.45\textwidth]{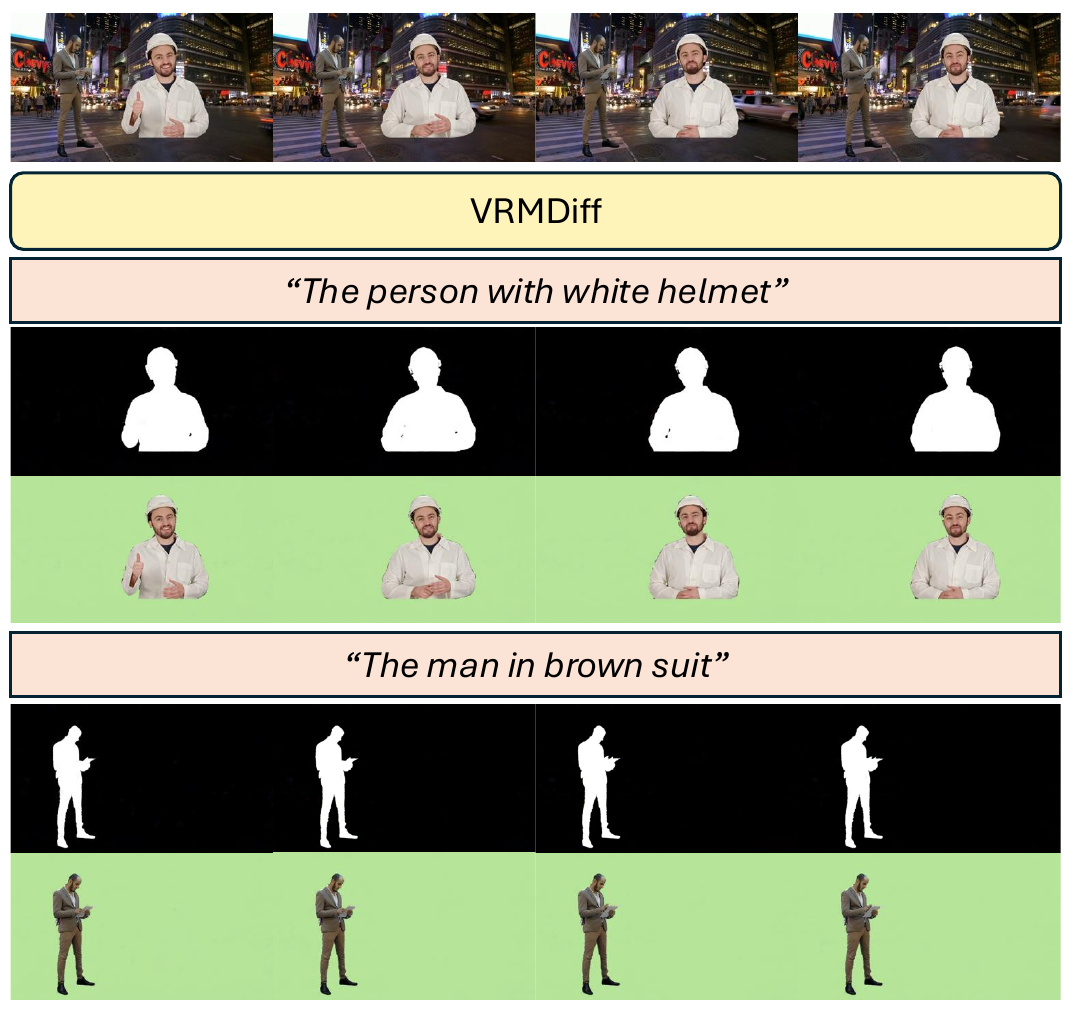}
    \caption{\textbf{Visualization results of referring matting method~\ourmethod.} We treat video matting as a generation task with referring capabilities. By inputting a caption that describes the instance, our model outputs the corresponding instance's matte.}
    \label{fig.teaser}
\end{figure}

Basically, in video referring matting, given a video and a caption describing an instance, we aim to output the matte corresponding to that instance specified.
This task not only requires understanding the semantic content of videos and referring expressions but also dealing with issues such as object occlusion, changing lighting conditions, and overlapping objects over time. 
Moreover, existing video matting datasets are limited in number and do not simultaneously provide captions, videos, and instance-level mattes, which obstructs the development of data-driven approaches for this problem.

In this paper, we introduce Video Referring Matting, a new task that combines the challenges of video matting and referring expression comprehension. 
To facilitate research in this field, we propose~\ourdataset{}, the first large-scale dataset for video referring matting.~\ourdataset{} contains 10,000 videos, each with up to five foreground objects composited onto dynamic backgrounds. 
Each object is associated with a high-quality alpha matte and descriptive captions generated using the Tarsier~\cite{wang2407tarsier} model, enabling fine-grained and natural language based object referencing.

To address the challenges and limitations in video referring matting outlined earlier, we propose a diffusion-based framework that effectively handles temporal dynamics and instance differentiation in videos. Diffusion models have demonstrated remarkable performance in image\cite{rombach2022high,podell2023sdxl,peebles2023scalable,zhang2023adding,chen2024anydoor,wang2024instancediffusion} and video generation \cite{yang2025cogvideox,jin2024pyramidal,ma2024latte,chen2024videocrafter2,guo2024animatediff,zhou2025storydiffusion,chen2024videocrafter2} tasks, as well as in multitask settings beyond RGB images and videos\cite{nguyen2023dataset,sheynin2024emu,geng2024instructdiffusion,qi2024unigs,ke2024repurposing}. Building on these advancements, our model extends the standard diffusion process by integrating video conditional inputs and referring expressions. This integration enables the generation of accurate and temporally consistent alpha mattes based on both visual and textual information, without relying on mask guidance. Our approach effectively addresses the difficulties in maintaining temporal consistency and correctly interpreting referring expressions in videos with complex dynamics and multiple overlapping objects. Moreover, the introduction of our comprehensive \ourdataset{} dataset facilitates the training of such models, enabling them to learn from a rich set of annotated videos and natural language descriptions.

One key challenge in our task is handling overlapping objects and ensuring that the model correctly interprets referring expressions to isolate the target object. 
To address this issue, we introduce a novel latent contrastive learning approach. 
Specifically, we design a Latent-InfoNCE loss function that operates in the latent space of a 3D variational autoencoder. 
This loss encourages the model to produce latent representations close to the ground truth of the correct object while pushing away representations of incorrect or overlapping objects. 
This strategy effectively mitigates ambiguity and enhances the model's ability to distinguish multiple objects based on referring expressions.

Our main contributions are summarized as follows:

\begin{itemize}
    \item Introduction of a new task and dataset: We define the video referring matting task and construct~\ourdataset ~dataset, a comprehensive dataset containing 10,000 videos, each with high-quality alpha mattes and corresponding natural language captions for each object.
    \item Diffusion-based matting framework: We develop a video diffusion model that integrates video conditional inputs and referring expressions, without relying on mask guidance.
    \item Latent contrastive learning: We propose a latent contrastive loss to enhance the model's ability to distinguish different objects in complex scenes, improving matting accuracy in multi-object and overlapping scenarios.
    \item Extensive experimental validation: We conduct thorough experiments to demonstrate the effectiveness of our method, showing significant improvements in matting quality and temporal coherence compared to baselines.
    
\end{itemize}

We believe our work lays the foundation for future research in video referring matting and opens up new possibilities for advanced applications in video processing and video understanding.

\section{Related Work}
\label{sec:related}

\noindent\textbf{Image and Video Matting.}
Image matting is a technique used to extract foreground objects from images by estimating their transparency, known as the alpha matte. Traditional image matting methods can be broadly categorized into two approaches: color sampling-based methods and alpha propagation-based methods. Color sampling-based methods~\cite{chuang2001bayesian,feng2016cluster,he2011global,aksoy2017designing,bai2007geodesic} leverage low-level features to differentiate the transition areas between the foreground and background, relying on the local smoothness assumption of image statistics. However, color sampling-based methods can suffer from the problem of matte discontinuities. To address this issue, propagation-based approaches~\cite{aksoy2017designing,chen2013knn,grady2005random,levin2007closed,sun2004poisson} aim to achieve optimal alpha values in unknown regions by leveraging the affinities of neighboring pixels. Recently, deep learning methods~\cite{li2021deep,xu2017deep,yu2021high,zhu2017fast,wang2018deep,cho2016natural,levin2007closed,shen2016deep} utilize the neural network to estimate the alpha matte with different forms of  auxiliary inputs such as trimaps~\cite{cai2019disentangled,li2020natural,xu2017deep}, sparse scribbles~\cite{levin2007closed}, background images~\cite{lin2021real}, coarse maps~\cite{yu2021mask} and mask guidance~\cite{liu2020boosting,yu2021mask}. In recent work, Referring Image Matting~(RIM)~\cite{li2023referring} has been proposed to extract the alpha matte of a specific object based on the provided textual descriptions. Matting Anything Model~\cite{li2024matting} adopt Segment Anything Model~\cite{kirillov2023segment} to perform the image matting.

In video matting, the accuracy of alpha matte estimation in a video sequence can be improved by exploiting the temporal context. Early works~\cite{apostoloff2004bayesian,choi2012video,li2013motion} in video matting merely extended image matting techniques to the temporal dimension but often resulted in unsatisfactory performance. Similar to image matting, trimaps are also utilized in trimap-based video matting methods~\cite{huang2023end,sun2021deep,zhang2021attention,seong2022one} to facilitate the fusion and alignment of temporal features through spatio-temporal feature aggregation. Apart from trimap-based methods, another line of works~\cite{lin2023adaptive,lin2021real,sengupta2020background,ke2022modnet,lin2022robust} adopt trimap-free approach to predict alpha mattes for the remaining video sequences. Recently, some studies~\cite{li2024vmformer,li2023videomatt,lin2023adaptive,cheng2022masked} have explored the use of attention mechanisms in video matting, achieving performance comparable to that of traditional CNN-based methods. Furthermore, to achieve high-quality results in video matting, Video Instance Matting~\cite{li2024video,huynh2024maggie} has been proposed to extract alpha mattes for individual instances separately.

\noindent\textbf{Diffusion Models for Generation.}
The diffusion model~\cite{rombach2022high} receive numerous attentions from the research community and extensive works~\cite{sauer2024fast,lu2024coarse,crowson2024scalable,ding2023patched} has been proposed to generate high-resolution images with text supervision by diffusion models. In recent years, substantial efforts have been devoted to video synthesis using diffusion models. Some early works~\cite{EsserCAGG23,ho2022video,Hong0ZL023,khachatryan2023text2video,wu2023tune} extend the text-to-image~(T2I) models to synthesize the videos. One potential solution~\cite{ge2023preserve,li2023videogen,SingerPH00ZHYAG23,blattmann2023align} is to modify the original T2I models by incorporating additional temporal modules and training the new models on video datasets. Apart from the text-to-video generation, some works~\cite{wang2023videocomposer,guo2024animatediff,hu2024animate} also explore the synthesis of the videos from the images. For instance, AnimateDiff~\cite{guo2024animatediff} generates videos by applying weighted mixing of image latents and random noise during the denoising process. However, these methods can't preserve the inherent temporal consistency in the video. To alleviate the heavy computational cost of training, some studies~\cite{shi2024bivdiff,Zhang0J0Z024,qi2023fatezero} have explored training-free methods for video generation. However, these training-free methods struggle to effectively preserve cross-frame consistency in textures and other details. To achieve fine-grained control over the details of synthesized videos, such as texture and motion patterns, some works~\cite{esser2023structure,wang2023videocomposer,xing2024make,wang2024recipe} leverage additional signals, such as depth or sketches, to guide the video generation process.

\noindent\textbf{Diffusion for Dense Prediction.}
Due to the remarkable success of diffusion models in generation tasks, there is a growing interest in incorporating the generative models into dense prediction tasks, such as depth estimation and semantic segmentation. Recently, some works~\cite{amit2021segdiff,chen2023diffusiondet,chen2023generalist,sheynin2024emu,geng2024instructdiffusion,qi2024unigs,wang2024semflow} attempted to perform the visual perception tasks using the diffusion models. These works~\cite{BaranchukVRKB22,xu2023open,zhao2023unleashing} either reuse or merge latent features of U-Nets~\cite{Ronneberger2015UNet} to perform segmentation and depth estimation tasks. Additionally, some efforts have been made to bridge the gap between the generation tasks and the dense prediction tasks. Lee et.al.~\cite{lee2024exploiting} leveraged the the pretrained text-to-image models as a prior and proposed an image-to-prediction diffusion process to adapt the text-to-image models for the dense prediction problems. He et.al.~\cite{he2024lotus} proposed a task switcher to preserve the fine-grained details during dense annotation generation.  Other approaches~\cite{ji2024dpbridge,BaranchukVRKB22,ji2023ddp} accomplish the same goal by formulating the dense prediction tasks as visual-condition-guided generation processes.

\section{Method}

\subsection{Video Referring Matting}
\subsubsection{Problem Definition}

\begin{table}[h]
    \centering
    \begin{tabular}{lcc}
        \toprule
        Task  & Instance & Text-guild \\ 
        \midrule
        Video Matting & \xmark & \xmark \\ 
        \hline
        Video Instance Matting & \cmark & \xmark \\ 
        \hline
        Video Referring Matting & \cmark & \cmark \\ 
        \bottomrule
    \end{tabular}
    \caption{Comparison between video referring matting and related video matting task.}
    \label{tab:tasks}
    \vspace{-2mm}
\end{table}

Unlike video matting and video instance matting, as shown in Table~\ref{tab:tasks}, our task not only distinguishes different instances but also uses captions as guidance to specify the object to be matted.
We consider a video sequence \( I \in \mathbb{R}^{T \times H \times W \times 3} \) consists of \( T \) frames, where each frame \( I_t \), with \( t \in \{1, \dots, T\} \), has spatial dimensions \( H \times W \). 
Given a natural language prompt \( P \) that describes a target object in the video, the goal of video referring matting task is to generate a temporally consistent alpha matte \( \alpha_t^i(P) \) that accurately isolates the described object while preserving fine-grained details such as transparency and soft boundaries. 
Unlike conventional video matting approaches, our model dynamically interprets both visual and textual information to infer precise object boundaries directly from the given prompt and video.

To achieve this process, our model learns to predict an alpha matte for each video frame, capturing fine-grained transparency and soft object boundaries while maintaining temporal consistency. 
Given a background image \( B_t \) and a foreground layer \( F_t \), the observed video frame can be formulated as:

\begin{equation}
    I_t = \alpha_t(P) \circ F_t + \left(1 - \alpha_t(P) \right) \circ B_t,
\end{equation}
where \( \alpha_t(P) \in [0,1]^{H \times W \times 1} \) represents the alpha matte of the target object in frame \( t \), conditioned on the prompt \( P \), ensuring that only the described region contributes to the final matte. The foreground layer \( F_t \in \mathbb{R}^{H \times W \times 3} \) represents the region of interest extracted from the input video, while \( B_t \in \mathbb{R}^{H \times W \times 3} \) represents the background. The operator \( \circ \) denotes the Hadamard product, enforcing per-element blending of the foreground and background.

Our approach utilizes a diffusion-based framework CogVideoX~\cite{yang2025cogvideox} to iteratively refine the predicted alpha matte, ensuring spatial precision and temporal consistency.
By conditioning the diffusion process on both the input video and the prompt, the model progressively denoises latent representations to generate high-quality mattes. 
This formulation allows the model to generalize across diverse objects, varying backgrounds, and complex motion patterns, making it well-suited for flexible and scalable video editing applications.

\subsubsection{Dataset}
\begin{figure}[tp]
    \centering
    \includegraphics[width=0.45\textwidth]{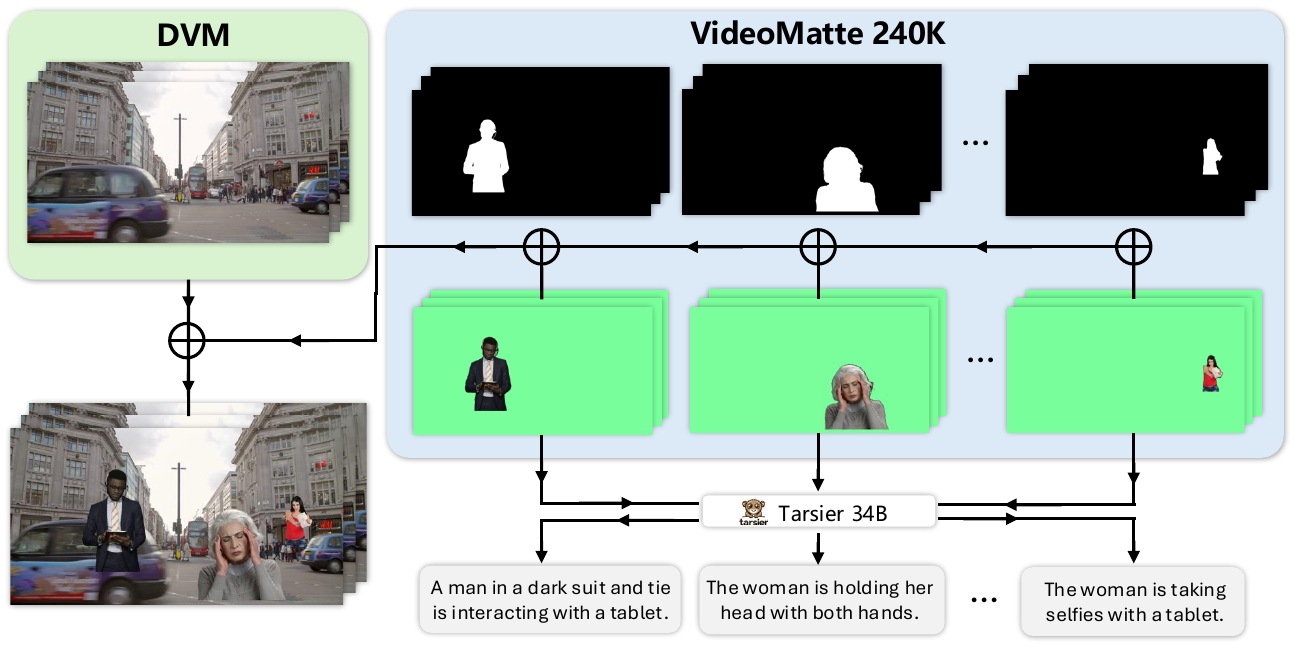}
    \caption{\textbf{Data pipeline.} For the background videos, we sample from DVM, and for the foreground instances, we sample from VideoMatte240K. The foreground instances are composited onto the background videos. The instance-level captions are generated by the vision-language model Tarsier, using the matte-extracted instance videos as input.}
    \label{fig.data_pipeline}
    \vspace{-5mm}
\end{figure}

Compared to image matting, video matting lacks sufficient datasets to support the training of diffusion models. 
Therefore, as the data pipeline shown in Figure~\ref{fig.data_pipeline}, we propose~\ourdataset, a multi-instance video matting dataset with high-quality alpha mattes and corresponding captions for each instance, totaling 10K videos of 120 frames each. 
Since the foregrounds in VideoMatte240K~\cite{lin2021real} do not include the original backgrounds, we sample background videos from DVM~\cite{sun2021deep} and foreground instances from VideoMatte240K. 
Specifically, we set the maximum number of foreground instances per video to five. 
For each randomly sampled background video, we determine the number of foreground instances to include by sampling from a normal distribution ranging from one to five. 
We then randomly select that number of foreground instances from VideoMatte240K. 
Each foreground instance is resized, repositioned, and applied onto the background video, the corresponding alpha matte undergoes the same transformation during this step.
Moreover, we limit the size differences among foreground instances within the same video to prevent significant disparities.
For the text captions, we use the 34B Tarsier~\cite{wang2407tarsier}, inputting the foreground instance videos to generate descriptive and diverse captions.

\begin{figure*}[tp]
    \centering
    \includegraphics[width=0.95\textwidth]{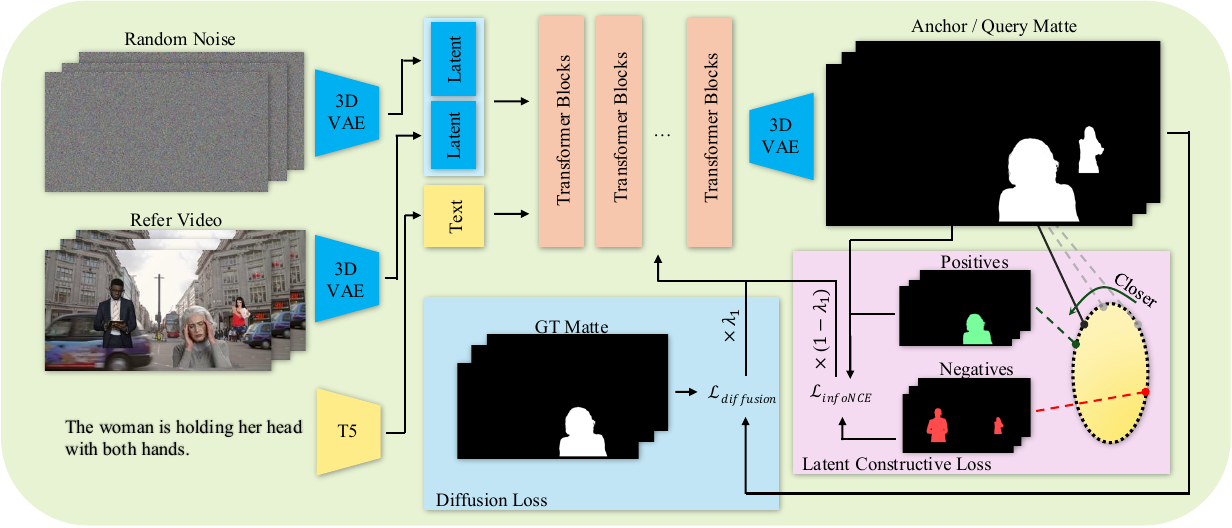}
    \caption{\textbf{Overview of~\ourmethod{} framework}. Similar to CogVideoX, our method performs denoising in the latent space using a 3D VAE. We input the conditional video and the corresponding referring caption, and the model is expected to output the alpha matte decoded into RGB space by the 3D VAE. Besides the diffusion loss, we employ a latent contrastive loss. During training, we use the model's latent output as the anchor, the ground truth matte as the positive sample, and instance mattes not corresponding to the caption as negative samples. This approach enhances the model's ability to distinguish instances and improves text-to-instance alignment.}
    \label{fig.framework}
    \vspace{-2mm}
\end{figure*}

\subsection{Diffusion Method}

In our video referring matting framework shown in Figure~\ref{fig.framework}, diffusion models serve as the core mechanism for refining predicted alpha mattes by progressively denoising a latent representation of the input video frames. The model begins with an initial noisy estimate and iteratively removes noise, ensuring the generated alpha mattes maintain both high fidelity and temporal consistency across frames.

To formalize this process, we define a forward diffusion model that gradually corrupts the original clean video frame \( x_0 \) by adding Gaussian noise at each step. The corrupted latent representation at time step \( t \), denoted as \( x_t \), follows the distribution:
\begin{equation}
    q(x_t | x_0) = \mathcal{N}(\sqrt{\bar{\alpha}_t} x_0, (1 - \bar{\alpha}_t) I),
\end{equation}
where \( q(x_t | x_0) \) represents the probability distribution of \( x_t \) given the clean frame \( x_0 \). The term \( \bar{\alpha}_t \) is a variance schedule that determines the amount of noise added at step \( t \), gradually increasing as \( t \) progresses. The identity matrix \( I \) ensures that the noise introduced at each step follows an isotropic Gaussian distribution.

To recover the clean video frame \( x_0 \), the reverse process aims to iteratively remove noise from \( x_t \) by predicting an intermediate latent representation at each step. The probability distribution of the denoised frame at step \( t-1 \), given \( x_t \), is formulated as:

\begin{equation}
    p_\theta(x_{t-1} | x_t) = \mathcal{N}(\mu_\theta(x_t, t), \sigma_t^2 I),
\end{equation}
where \( p_\theta(x_{t-1} | x_t) \) represents the probability distribution of the denoised latent state \( x_{t-1} \) given the current noisy state \( x_t \). The term \( \mu_\theta(x_t, t) \) is a function that predicts the mean of this distribution, parameterized by \( \theta \), which represents the learnable weights of the denoising model. The variance \( \sigma_t^2 \) controls the uncertainty of the prediction at each step.

By progressively refining \( x_t \), the model is able to recover high-quality alpha mattes that align with the textual prompt while ensuring smooth transitions between video frames.

\subsubsection{Video-Conditioned Diffusion}

Unlike standard diffusion-based video generation models CogVideoX~\cite{yang2025cogvideox}, where the diffusion process is guided solely by text prompts and temporal dependencies, our framework introduces an additional video-conditioned input.
This enables the model to leverage both spatiotemporal coherence from the reference video and the generative capabilities of diffusion models.
Specifically, we expand the input channel dimension from 16 to 32 to accommodate the additional video signal.

Given an input reference video \( x_t^{\text{video}} \), we encode it using a 3D Variational Autoencoder (3D VAE) to obtain its latent representation \( z_t^{\text{video}} \). Similarly, a pure Gaussian noise sample \( x_t^{\text{noise}} \) is encoded using the same 3D VAE, producing \( z_t^{\text{noise}} \). These two latent representations are then concatenated to form the input for the diffusion transformer:

\begin{equation}
    z_t = \text{Concat}(z_t^{\text{video}}, z_t^{\text{noise}}),
\end{equation}
where \( z_t \) represents the latent representation at time step \( t \), composed of:
 \( z_t^{\text{video}} \), the latent embedding obtained from the encoded reference video.
 \( z_t^{\text{noise}} \), the latent representation of the noise injected during the forward diffusion process.

This concatenated latent representation is then fed into the diffusion transformer, allowing the model to maintain temporal consistency across frames while benefiting from the generative flexibility of the diffusion process.

\subsubsection{Text-Conditioned Matte Generation}

Beyond video conditioning, our model incorporates textual guidance to enhance the semantic alignment between the generated matte and the user-provided prompt. Instead of dynamically injecting text information at each transformer block, we encode the textual description once and integrate it into the initial input representation. Given an input prompt \( T \), we obtain its dense embedding using a T5 encoder:
\begin{equation}
    e_{\text{text}} = \text{T5-Encoder}(T).
\end{equation}

The diffusion transformer then receives a combination of the video-conditioned latent representation and the text embedding as input:
\begin{equation}
    h_0 = \text{TransformerInput}(z_t, e_{\text{text}}),
\end{equation}
where \( h_0 \) is the initial feature embedding. The term \( z_t \) represents the concatenated latent representation of the video-conditioned embedding and the noise input.

Once injected into the transformer, the feature representation is iteratively refined through multiple layers. The updated representation at layer \( l \) is given by: \( h_l = \text{TransformerBlock}(h_{l-1}) \)   where \( h_l \) denotes the feature representation at layer \( l \) of the diffusion transformer.

\subsection{Latent Contrastive Learning}

Referring matting presents a fundamental challenge: different objects in the scene may share similar visual features, making it difficult to differentiate the correct foreground object based solely on visual cues. 
Our goal is to ensure that the predicted matte aligns with the reference description while maintaining separation between ambiguous objects. To achieve this, we leverage contrastive learning in the latent space, encouraging the model to generate mattes that are not only structurally accurate but also well-aligned with the text prompt.

A direct approach to training the model would be to minimize the pixel-wise L2 loss:
\begin{equation}
\mathcal{L}_{\text{pixel}} = \| M - \hat{M} \|^2.
\end{equation}

However, this approach fails to capture high-level object semantics and struggles with ambiguous cases where multiple objects share similar textures. To explicitly enforce separation in the latent space, we assume that the predicted matte embedding \( \hat{z} \) should be closer to the true matte embedding \( z \) than to incorrect matte embeddings. This motivates the use of contrastive learning, where we minimize the distance between positive pairs while pushing negative samples apart:

\begin{equation}
\mathcal{L}_{\text{InfoNCE}} = -\log \frac{\exp(\text{sim}(\hat{z}, z^+)/\tau)}{\sum_{z^-} \exp(\text{sim}(\hat{z}, z^-)/\tau)},
\end{equation}
where \( z^+ = \text{VAE}(M) \) is the ground truth matte embedding, \( z^- \) represents negative samples, i.e., incorrect matte embeddings, \( \text{sim}(a, b) \) is the cosine similarity function:
  \begin{equation}
  \text{sim}(a, b) = \frac{a \cdot b}{\|a\| \|b\|},
  \end{equation}
and \( \tau \) is a temperature scaling parameter.

While contrastive learning in pixel space is possible, it is computationally inefficient and fails to capture object-level relationships. Instead, we perform contrastive learning in the latent space by leveraging the 3D VAE, which achieves an \(8 \times 8 \times 4\) compression from pixels to latent representations, thereby significantly reducing computational complexity. In our framework, we define \(p(z)\) as the ground truth latent distribution obtained by encoding the ground truth matte \(M\) via the VAE (i.e., \(p(z)=\text{VAE}(M)\)), and \(q(\hat{z})\) as the predicted latent distribution obtained from the predicted matte \(\hat{M}\) (i.e., \(q(\hat{z})=\text{VAE}(\hat{M})\)). To quantify the discrepancy between these distributions, we consider the Kullback-Leibler (KL) divergence:
\[
D_{\text{KL}}(q(\hat{z}) \| p(z)) = \int q(\hat{z}) \log \frac{q(\hat{z})}{p(z)} \, d\hat{z},
\]
which measures how well the predicted latent distribution \(q(\hat{z})\) approximates the ground truth distribution \(p(z)\). This formulation encourages the model to generate latent representations that align closely with the semantics of the ground truth matte while filtering out high-frequency noise. Then, using pixel space representations results in sparsely distributed embeddings, making KL divergence estimation unreliable. Formally, the variance of the KL estimate can be written as:
\begin{equation}
\text{Var}(D_{\text{KL}}) = \mathbb{E}[(\log q(\hat{z}) - \log p(z))^2] - D_{\text{KL}}^2.
\end{equation}

Since pixel space embeddings are high-dimensional and sparse, the expectation term grows large, leading to an unstable estimation of \( D_{\text{KL}} \). In contrast, by mapping the input to a structured latent space, we obtain a more compact distribution, improving the stability of contrastive learning.

Another advantage of working in the latent space is its ability to absorb minor misalignments, increasing the tolerance to small prediction errors. Let \( M \in \mathbb{R}^{H \times W} \) denote the ground truth alpha matte, representing the per-pixel opacity of the target object, where \( M(i, j) \in [0,1] \) at each spatial location \( (i, j) \). Suppose the predicted matte \( \hat{M} \) contains minor noise \( \epsilon \) in pixel space:
\begin{equation}
\hat{M} = M + \epsilon.
\end{equation}

Applying the VAE encoder to the noisy matte produces the corresponding latent representation:
\begin{equation}
\hat{z} = \text{VAE}(M + \epsilon).
\end{equation}

If the VAE is trained to minimize reconstruction loss, then the expected reconstruction error satisfies:
\begin{equation}
\mathbb{E} \|\text{VAE}^{-1}(z) - M\|^2 \leq \lambda \|\epsilon\|^2,
\end{equation}
where \( \lambda \) is a compression factor determined by the VAE architecture. Since \( \lambda < 1 \) in a well-trained VAE, the transformation effectively suppresses high-frequency noise, leading to a more robust matte representation. Consequently, small spatial errors in pixel space have a reduced impact in the latent space, resulting in more stable and temporally consistent matte predictions.

To jointly optimize matte reconstruction and feature separation, we define the final loss function:
\begin{equation}
\mathcal{L} = \lambda_1 \mathcal{L}_{\text{diffusion}} + (1 - \lambda_1) \mathcal{L}_{\text{InfoNCE}}.
\end{equation}
where \( \mathcal{L}_{\text{diffusion}} = \mathbb{E} \|\epsilon - \epsilon_\theta(z_t, t)\|^2 \) is the standard diffusion loss, and \( \mathcal{L}_{\text{InfoNCE}} \) ensures feature separation in latent space.

By applying contrastive learning in the latent space, our model improves the robustness of matte generation, achieving better alignment between textual descriptions and object-level representations.

\section{Experiments}

\begin{table}[t]
    \centering
    \begin{tabular}{lcccc}
        \toprule
        & MAD $(\downarrow)$   & MSE $(\downarrow)$    & Grad $(\downarrow)$ & Conn $(\uparrow)$ \\
        \midrule
        Baseline & 0.0875 & 0.0831 & \textbf{0.0012} & 0.9137 \\
        VRMDiff & \textbf{0.0689} & \textbf{0.0668} & 0.0014 & \textbf{0.9293} \\
        \bottomrule
    \end{tabular}
    \caption{\textbf{Evaluation of distance metrics for referring matting.} MAD, MSE, Grad and Conn stand for mean absolute differences, mean squared error, gradient and connectivity respectively.}
    \label{tab:video_referring_metric}
\end{table}

\begin{table}[t]
    \centering
    \begin{tabular}{lcccc}
        \toprule
        & RQ $(\uparrow)$     & TQ $(\uparrow)$    & MQ $(\uparrow)$     & VIMQ $(\uparrow)$ \\
        \midrule
        Baseline & 16.18 & 38.31 & 33.75 & 2.09 \\
        VRMDiff     & \textbf{37.00} & \textbf{74.59} & \textbf{49.66} & \textbf{13.71} \\
        \bottomrule
    \end{tabular}
    \caption{\textbf{Evaluation of matting quality metrics for referring matting.} RQ, TQ, MQ and VIMQ stand for recognition quality, tracking quality, instance-level matting quality and video instance-aware matting quality respectively.}
    \label{tab:video_referring_quality}
    \vspace{-2mm}
\end{table}

\begin{figure*}[t]
    \centering
    \includegraphics[width=0.99\textwidth]{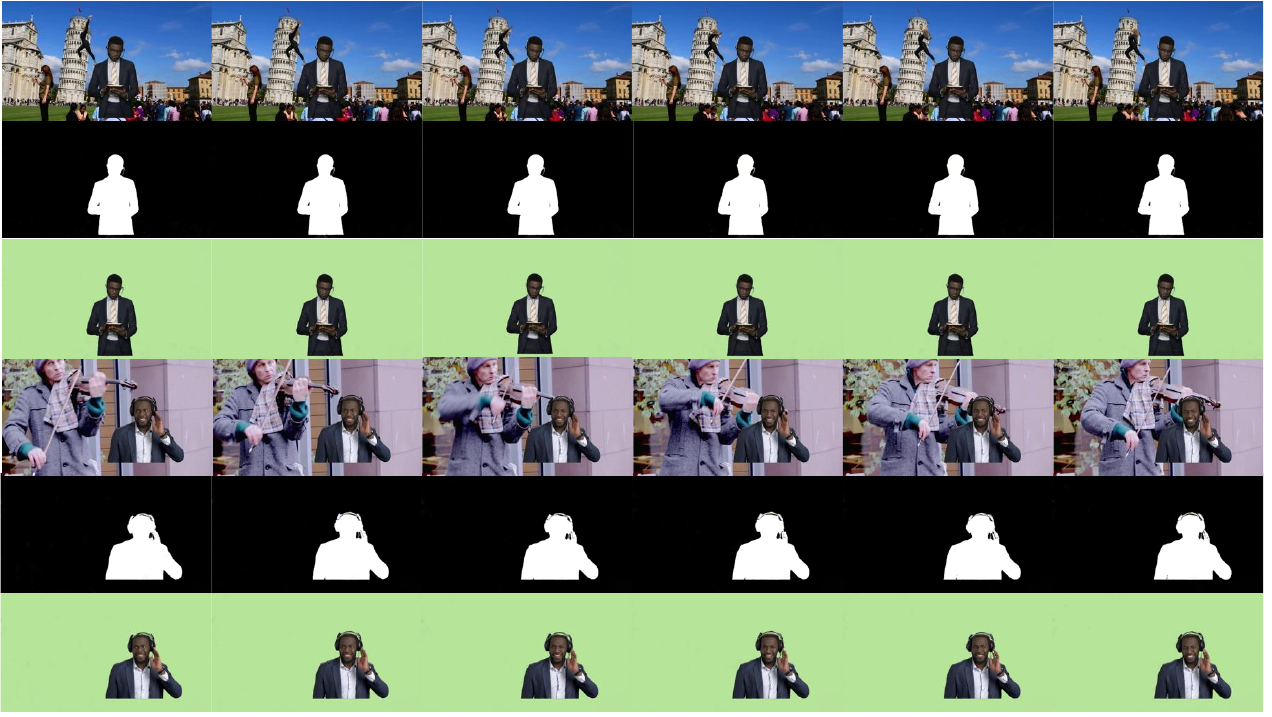}
    \vspace{-2mm}
    \caption{\textbf{Single instance matting qualitative results on~\ourmethod}. Six frames are evenly sampled from the video, with the horizontal axis representing time and the frame index gradually increasing. From top to bottom, the sequence is the input video, the output alpha matte, and the extracted instance obtained by applying the matte to the video. The prompts are \textit{A man in a dark suit and tie} and \textit{The man is wearing a dark blue suit jacket, a white shirt, and black headphones}.}
    \label{fig.single}
\end{figure*}

\begin{table*}[t]
    \centering
    \begin{tabular}{lcccccc}
        \toprule
        Method & Type & Backbone & MAD $(\downarrow)$   & MSE $(\downarrow)$    & Grad $(\downarrow)$ & Conn $(\uparrow)$ \\
        \midrule
        \cmidrule(lr){1-7}

        RVM~\cite{lin2022robust} & Instance-agnostic & MobileNet V3 & 0.0294 & 0.0258 & 0.0024 & 0.9609   \\
        RVM~\cite{lin2022robust} & Instance-agnostic & ResNet 50 & 0.0334 & 0.0294 & 0.0024 & 0.9549 \\

        \cmidrule(lr){1-7}
        VRMDiff (Ours) & Referring & CogVideoX 2B & 0.0689 & 0.0668 & 0.0014 & 0.9293 \\
        \bottomrule
    \end{tabular}
    \vspace{-2mm}
    \caption{\textbf{Comparison between instance and caption agnostic method and referring method.} MAD, MSE, Grad and Conn stand for mean absolute differences, mean squared error, gradient and connectivity respectively.}
    \label{tab:method_compare}
    \vspace{-2mm}
\end{table*}

In this section, we first evaluate our method's performance on the new video referring matting task using datasets involving multiple objects, and then compare it with previous instance and caption agnostic video matting methods. 
We use basic metrics such as MAD, MSE, Gradient, and Connectivity for general measurements, and also employ metrics focused on instance matting~\cite{li2024video}, including Matting Quality (MQ), Tracking Quality (TQ), Recognition Quality (RQ), and their product, Video Instance-aware Matting Quality (VIMQ), to better assess the quality of matting and the alignment between the referring condition and the generated results. 
In the ablation study, we investigate the capability of InfoNCE in latent space for contrastive learning, verifying the effectiveness of Latent InfoNCE.

\subsection{Experimental Settings}
We conduct our experiments using the~\ourdataset{} dataset synthesized from DVM and VideoMatte240K. 
This dataset includes 9000 training samples and 1000 validation samples. 
The training and validation samples are independently partitioned during dataset creation, derived from entirely different splits of DVM and VideoMatte240K. 
We initialize our model weights with the CogVideoX 2B text-to-video model. 
The weights of the original input channels are resampled to obtain weights for the video condition channels. 
Our experiments are conducted on two NVIDIA A40 48G GPUs, with a batch size of 1 per GPU, fine-tuning for 10000 steps using a learning rate of 1e-5.

\subsection{Video Referring Matting}

\begin{figure*}[t]
    \centering
    \includegraphics[width=0.99\textwidth]{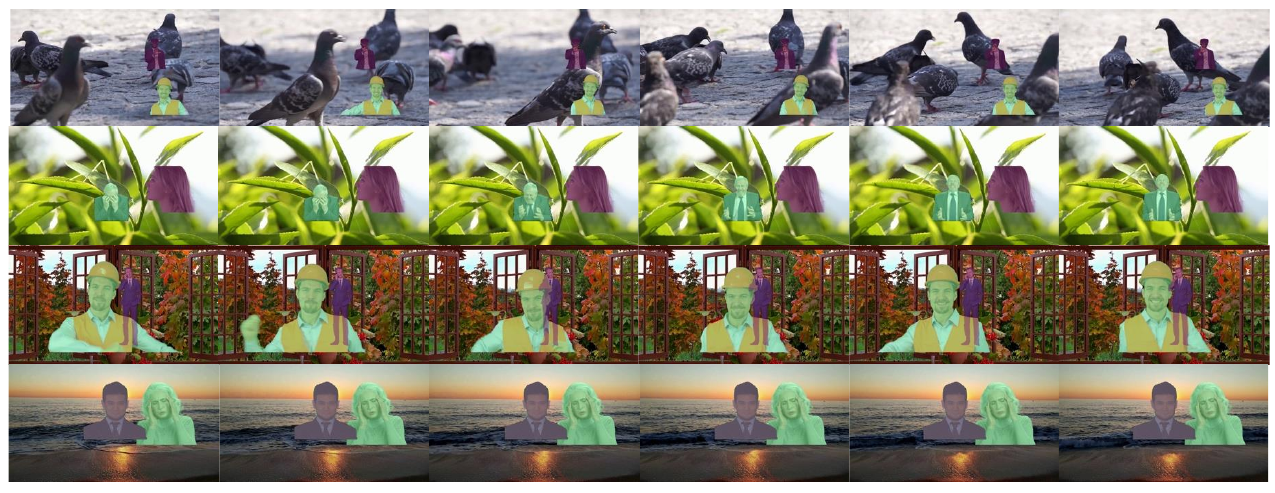}
    \vspace{-2mm}
    \caption{\textbf{Multiple instance matting qualitative results on~\ourmethod}. Six frames are evenly sampled from the video, with the horizontal axis representing time and the frame index gradually increasing. We use masks of different colors to represent the alpha mattes of different instances.}
    \label{fig.multi}
    \vspace{-3mm}
\end{figure*}

For the video referring matting task, since we are the first to propose this task, there are no prior works for direct comparison. 
Therefore, in our experiments, we compare our complete method with a baseline model that incorporates video conditioning into CogVideoX. 
All models are trained under the same conditions and data for the same number of steps. 
As shown in Table~\ref{tab:video_referring_metric} and~\ref{tab:video_referring_quality}, our method demonstrates improvements over the video-conditioned CogVideoX across various metrics, confirming the effectiveness of our approach. 
Additionally, our method exhibits better text-to-matte alignment and captures more continuous and consistent motion.

We compared instance-agnostic and caption-agnostic methods. 
These methods can only distinguish between foreground and background, unable to differentiate among different instances, and do not use captions as conditions. 
For these methods, we evaluated their performance on the validation set of our~\ourdataset{}, combining the instance matte ground truths into a single foreground matte. 
As shown in Table~\ref{tab:method_compare}, our class and caption aware method performs slightly lower than the class-agnostic method RVM in MAD, MSE, and Connectivity metrics but outperforms them in the Gradient metric. 
Although this comparison is relatively unfavorable to our method due to the more challenging task and objectives, achieving comparable performance demonstrates the effectiveness of our approach.

\subsection{Qualitative Results}
Figure~\ref{fig.single} illustrates our video matting generation capability in scenarios with complex backgrounds and single foregrounds. 
The referring ability not only accurately locates the instance specified by the caption in complex scenes, but also enables our generated mattes to extract high quality instances without including background elements precisely. 
Meanwhile, in Figure~\ref{fig.multi}, we present the quality of multi-object referring matting in situations with multiple foreground subjects under both simple and complex backgrounds. 
Our model maintains good text-to-instance alignment in these cases and performs well on both large-size and small-size objects.

\subsection{Ablation Study}

\begin{table}[t]
    \centering
    \begin{tabular}{lcccc}
        \toprule
        $\lambda$ & MAD $(\downarrow)$   & MSE $(\downarrow)$    & Grad $(\downarrow)$ & Conn $(\uparrow)$ \\
        \midrule
        0.1 & \textbf{0.0689} & \textbf{0.0668} & 0.0014 & \textbf{0.9293} \\
        0.25 & 0.0754 & 0.0721 & 0.0013 & 0.9240 \\
        0.5 & 0.0886 & 0.0824 & \textbf{0.0012} & 0.9138 \\
        \bottomrule
    \end{tabular}
    \caption{\textbf{Ablation study on contrastive loss weight.} RQ, TQ, MQ and VIMQ stand for recognition quality, tracking quality, instance-level matting quality and video instance-aware matting quality respectively.}
    \label{tab:ab_part1}
    \vspace{-2mm}
\end{table}

\begin{table}[t]
    \centering
    \begin{tabular}{lcccc}
        \toprule
        $\lambda$ & RQ $(\uparrow)$     & TQ $(\uparrow)$    & MQ $(\uparrow)$     & VIMQ $(\uparrow)$ \\
        \midrule
        0.1     & \textbf{37.00} & \textbf{74.59} & \textbf{49.66} & \textbf{13.71} \\
        0.25 & 25.62 & 53.31 & 38.28 & 5.23 \\
        0.5 & 7.41 & 18.59 & 9.68 & 0.13 \\
        \bottomrule
    \end{tabular}
    \caption{\textbf{Ablation study on contrastive loss weight.} MAD, MSE, Grad and Conn stand for mean absolute differences, mean squared error, gradient and connectivity respectively.}
    \label{tab:ab_part2}
    \vspace{-2mm}
\end{table}

The primary training objective of the diffusion model remains generation. 
We treat the video matting task as a video generation task and use contrastive loss to regulate text-to-instance alignment. 
Here, we ablate the weight lambda of the contrastive loss in the total loss. The results in Table~\ref{tab:ab_part1} and~\ref{tab:ab_part2} show that our choice of 0.1 achieves optimal performance on the majority of metrics.

\section{Conclusion}

In this paper, we introduces Video Referring Matting, a novel task that aims to extract the alpha matte of target objects in videos based on natural language queries. 
Treating the alpha matte as an RGB video and using diffusion models for generative matting, integrates video conditional inputs and referring expressions without relying on mask guidance.
We propose a latent contrastive learning approach that improves matting accuracy and temporal coherence, enhance the model's ability to distinguish different objects in complex scenes.
Additionally, we present a large-scale captioned video referring matting dataset.
Extensive experiments demonstrate the effectiveness of our method, showing significant improvements over baseline approaches.

\newpage
{
    \small
    \bibliographystyle{ieeenat_fullname}
    \bibliography{main}
}


\appendix
\section{Appendix}

This supplementary material provides additional visualization results, details about the ~\ourdataset~ video referring matting dataset.
The contents of this supplementary material are organized as follows:

\begin{itemize}
    \item \ourdataset~Dataset details.
    \item Additional visualization results, including examples of video referring matting.
\end{itemize}

Additionally, please see the mp4 file in our supplementary material to view the \underline{\textit{recorded video}} that provides a concise overview of our paper.

\section{\ourdataset~Dataset Details}

\begin{figure}[h]
    \centering
    \includegraphics[width=0.45\textwidth]{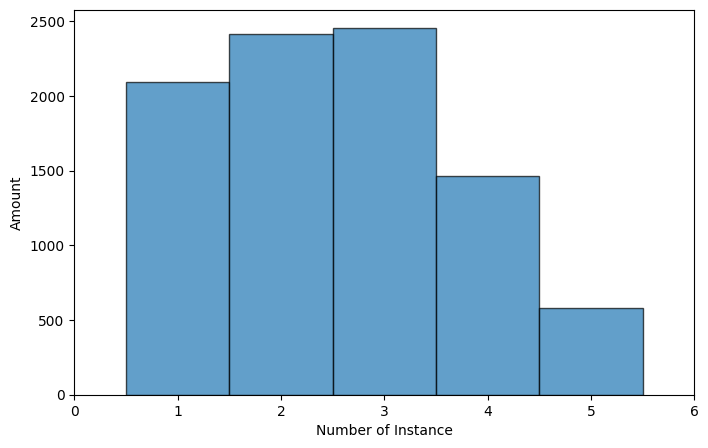}
    \caption{\textbf{Distribution of instance number in the video referring matting dataset.} The X-axis represents the groups of the total number of instances in a single video, while the Y-axis represents the total number of videos appearing in this group.}
    \label{fig.supp_data}
\end{figure}

We have recorded that in the 9,000 samples of the training set in our VRM-10K dataset, the mean number of instances is 2.5581 with a standard deviation of 1.1926. 
The overall distribution is shown in Figure~\ref{fig.supp_data}.

\section{More Visualization Results}
\begin{figure*}[ht]
    \centering
    \includegraphics[width=0.9\textwidth]{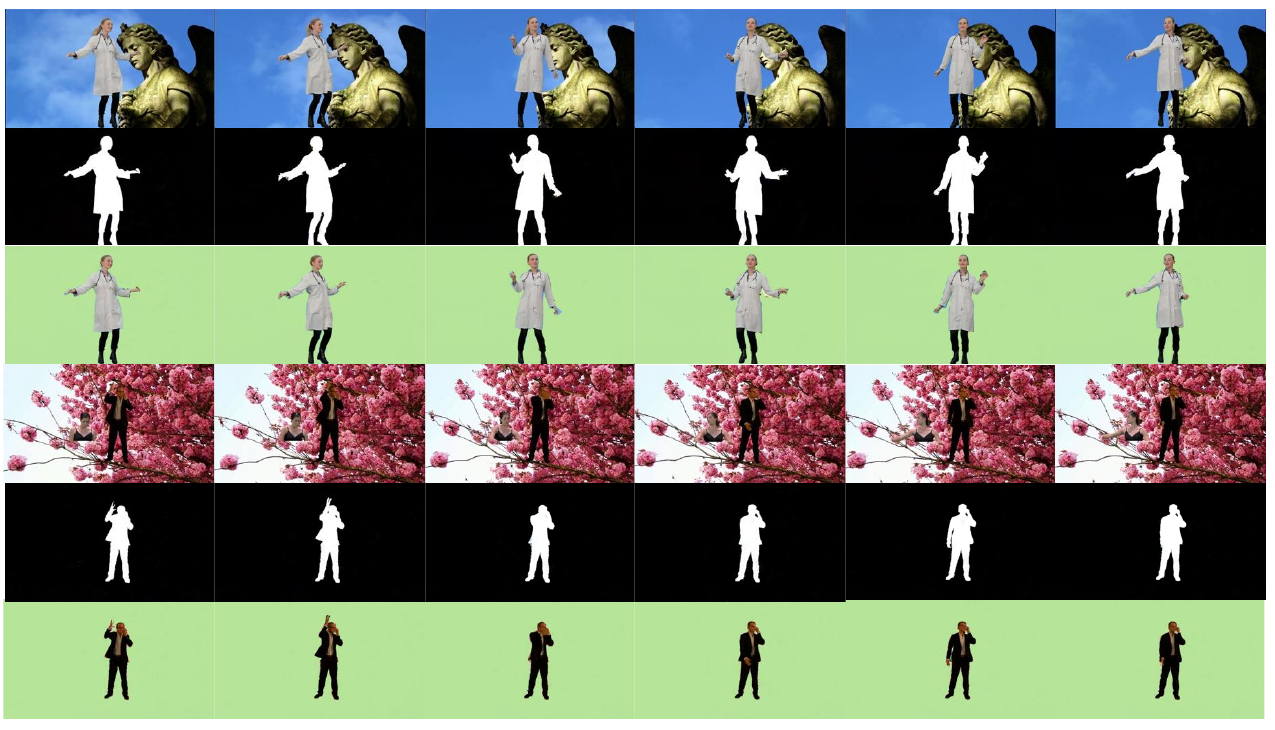}
    \caption{\textbf{More examples on referring video matting quality}. Six frames are evenly sampled from the video, with the horizontal axis representing time and the frame index gradually increasing. From top to bottom, the sequence is the input video, the output alpha matte, and the extracted instance obtained by applying the matte to the video.}
    \label{fig.supp1}
\end{figure*}

\begin{figure*}[ht]
    \centering
    \includegraphics[width=0.9\textwidth]{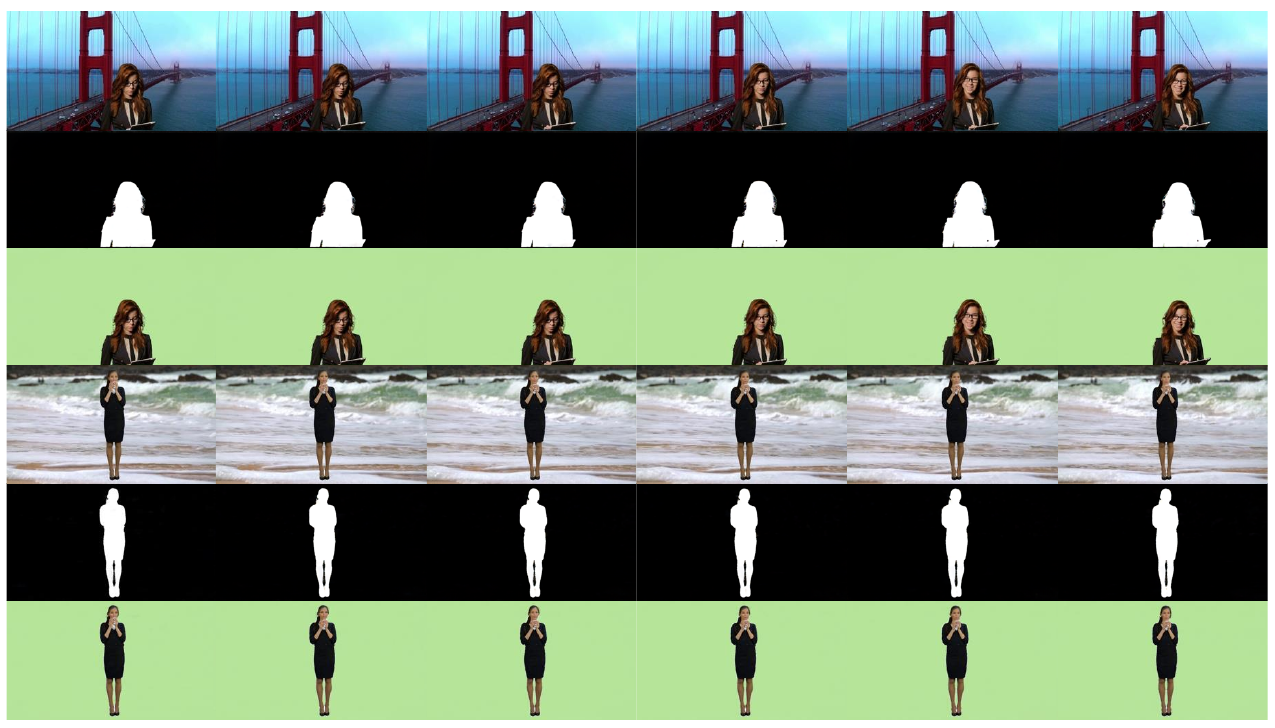}
    \caption{\textbf{More examples on referring video matting quality}. Six frames are evenly sampled from the video, with the horizontal axis representing time and the frame index gradually increasing. From top to bottom, the sequence is the input video, the output alpha matte, and the extracted instance obtained by applying the matte to the video.}
    \label{fig.supp2}
\end{figure*}

We provide additional visualizations in Figures~\ref{fig.supp1} and~\ref{fig.supp2}.
In various scenarios with both simple and complex backgrounds, our referring matting method consistently achieves precise alpha mattes.

\end{document}